\documentclass[a4paper,fleqn]{cas-dc}

\usepackage[authoryear]{natbib}
\usepackage{caption}
\usepackage{makecell}
\usepackage{subcaption}

\def\tsc#1{\csdef{#1}{\textsc{\lowercase{#1}}\xspace}}
\tsc{WGM}
\tsc{QE}
\tsc{EP}
\tsc{PMS}
\tsc{BEC}
\tsc{DE}

\begin{document}
\let\WriteBookmarks\relax
\def\floatpagepagefraction{1}
\def\textpagefraction{.001}

\shorttitle{Multitask learning for recognizing stress and depression in social media}

\shortauthors{L.Ilias, D.Askounis}  

\title [mode = title]{Multitask learning for recognizing stress and depression in social media}  



%

\author[1]{Loukas Ilias}[orcid=0000-0002-4483-4264]

\cormark[1]


\ead{lilias@epu.ntua.gr}


\credit{Conceptualization, Methodology, Software, Validation, Formal Analysis, Visualization, Investigation, Data Curation, Writing - Original Draft, Writing - Review \& Editing}

\affiliation[1]{organization={Decision Support Systems Laboratory, School of Electrical and Computer Engineering, National Technical University of Athens},
            addressline={Zografou}, 
            city={Athens},
            postcode={15780}, 
            country={Greece}}

\author[1]{Dimitris Askounis}[orcid=0000-0002-2618-5715
]


\ead{askous@epu.ntua.gr}


\credit{Supervision, Project Administration, Writing - Review \& Editing}



\cortext[1]{Corresponding author}

\begin{abstract}
Stress and depression are prevalent nowadays across people of all ages due to the quick paces of life. People use social media to express their feelings. Thus, social media constitute a valuable form of information for the early recognition of stress and depression. Although many research works have been introduced targeting the early recognition of stress and depression, there are still limitations. There have been proposed multi-task learning settings, which use depression and emotion (or figurative language) as the primary and auxiliary tasks respectively. However, although stress is inextricably linked with depression, researchers face these two tasks as two separate tasks. To address these limitations, we present the first study, which exploits two different datasets collected under different conditions, and introduce two multitask learning frameworks, which use depression and stress as the main and auxiliary tasks respectively. Specifically, we use a depression dataset and a stressful dataset including stressful posts from ten subreddits of five domains. In terms of the first approach, each post passes through a shared BERT layer, which is updated by both tasks. Next, two separate BERT encoder layers are exploited, which are updated by each task separately. Regarding the second approach, it consists of shared and task-specific layers weighted by attention fusion networks. We conduct a series of experiments and compare our approaches with existing research initiatives, single-task learning, and transfer learning. Experiments show multiple advantages of our approaches over state-of-the-art ones.
\end{abstract}




\maketitle

\section{Introduction}

The World Health Organization (WHO)\footnote{https://www.who.int/news-room/fact-sheets/detail/depression} states that around 280 million people in the world suffer from depression. Specifically, depression comes with a loss of pleasure or interest for a long period of time. According to the study\footnote{https://my.clevelandclinic.org/health/diseases/9290-depression}, many factors contribute to depression, including brain chemistry, genetics, stressful life events, medical conditions, and side effects of medications. In terms of stressful events, the authors mention that difficult experiences, such as the death of a person, divorce, isolation and lack of support, may cause depression. According to the WHO\footnote{https://www.who.int/news-room/questions-and-answers/item/stress}, stressful situations can cause anxiety and depression. Therefore, stress and depression are relevant situations. Social media can be proven a valuable form of information and knowledge extraction, since people use social media to share their feelings.

There have been proposed many studies for recognizing stress and depression in social media posts. However, the majority of these approaches train these tasks separately by introducing in this way single-task learning models \citep{info:doi/10.2196/28754,10154134}. Other studies \citep{8681445} propose feature extraction strategies and train traditional machine learning classifiers, including Logistic Regression (LR), Random Forests (RF), Decision Trees (DT), etc. Employing feature extraction approaches is a time consuming process, since it demands domain expertise. Thus, the optimal set of features may not be found. Recently, there have been proposed studies \citep{8910655,9141494} exploiting clinical interviews, which employ multi-task learning frameworks by defining as the primary task the identification of depression and as the auxiliary task the emotion classification. However, little work has been done in terms of detecting stress or depression in social media in a multi-task learning setting. Although stress and depression are linked, only a limited number of studies \citep{10068655} has experimented with identifying these mental conditions in a multitask learning setting. However, these studies utilize one single dataset and reformulate the multilabel problem into a multi-task learning setting.

To tackle these limitations, we present the first study, which detects depressive and stressful posts at the same time in a multi-task learning framework by exploiting one depression dataset and one stressful dataset, which have been collected and labelled under different conditions. The multi-task learning consists of two tasks, namely the primary task and the auxiliary one. We define the depression detection task as the primary task, while the stress detection task is defined as the auxiliary one. Specifically, we propose two multi-task learning architectures. In terms of the first architecture, we pass each depressive and stressful post through a shared BERT layer. The outputs of the shared BERT layer are passed through two separate task-specific BERT layers, which are updated by each task separately. Regarding the second approach, we utilize both shared and task-specific layers and exploit also an attention fusion network to let the network decide by itself the weights of the task-specific and shared representations in the decision process. In terms of both architectures, both tasks are optimized jointly. In contrast with other studies \citep{10068655}, which use one single dataset and redesign the multilabel problem as a multi-task classification where one post indicates both depression and anxiety, we exploit two datasets, i.e., the one dataset includes depressive and non-depressive posts, while the other one includes stressful and non-stressful posts. These datasets have been collected and labelled under different conditions. At the same time, contrary to other studies, which employ only the anxiety \citep{10068655} as a task in a multi-task learning scenario, we exploit the Dreaddit dataset \citep{turcan-mckeown-2019-dreaddit} for recognizing stressful texts as an auxiliary task, which includes ten subreddits from five domains. Specifically, texts are pertinent to domestic violence, survivors of abuse, anxiety, stress, almost homeless, assistance, homeless, food pantry, relationships, and Post-Traumatic Stress Disorder (PTSD). For the reasons mentioned above, this study aims to recognize depressive and stressful posts in a more challenging way proving in this manner better generalizability of the introduced approaches. We conduct an exhaustive series of experiments and compare our introduced approaches with state-of-the-art approaches, single-task learning models, and transfer learning, i.e., we train single-task learning models on one task (stress) and fine-tune it on the other task (depression) and vice versa. Findings suggest that our introduced approaches are capable of achieving notable performance gains in comparison with the aforementioned baselines.

Our main contributions can be summarized as follows:

\begin{itemize}
    \item We predict stressful and depressive posts in a multitask learning setting utilizing two different datasets collected and labelled under different situations for defining the primary and auxiliary tasks.
    \item We exploit shared, task-specific layers, and attention fusion networks. To the best of our knowledge, this is the first study utilizing attention fusion networks for predicting stress and depression through social media posts.
    \item We compare our approaches with single-task learning models, transfer-learning strategies, and state-of-the-art approaches.
\end{itemize}

\section{Related Work}

Several multi-task learning architectures have been proposed for recognizing stress and depression in social media. The authors in \citet{ZHOU2021102119} introduced a multi-task learning framework, where the main task aims at identifying depressive posts (binary classification task), while the auxiliary task corresponds to the domain category classification task. The authors exploited an hierarchical neural network consisting of a sentence encoder and a document encoder. Also, the authors enhanced BiGRUs with topic information. Findings indicated that the conjunction of both the multi-task learning framework and the topic features achieved the best results. In \citet{WANG2022727}, the authors employed a multitask learning setting with multimodal inputs for predicting online depressed users in Weibo. In terms of the multi-task learning framework, the authors defined the manually extracted statistical feature classification task as the main task, while the word vector classification task was defined as the auxiliary one. In \cite{ghosh2022multitask}, the authors introduced a multitask learning approach consisting of the primary task, i.e., emotion recognition, and the secondary tasks, i.e., depression detection and sentiment classification. The authors introduced some architectures consisting of GloVe embeddings, BiGRU, Attention, and fully connected layers. Results showed that the secondary tasks improve the performance of the primary task. On the contrary, decrease in performance is noticed in case of single-task learning approaches. A multimodal multitask learning framework was proposed by \cite{9567734}, where the primary task was defined as a depression detection task, while the auxiliary task was defined as an emotion recognition task. The proposed deep neural network receives as inputs the user description field, the image profile, and some features extracted by the Natural Language Understanding tool of IBM Watson. The architecture exploits GloVe embeddings and comprises BiGRU, Attention, CNN, and fully connected layers. Findings suggested that the proposed multitask learning approach outperformed several single task learning baselines. In \citet{turcan-etal-2021-emotion}, the authors experimented with three ways of integrating emotion information into stress detection models, which can be categorized into multi-task learning and finetuning. The authors used the dataset introduced in \citet{turcan-mckeown-2019-dreaddit} for the task of stress detection, while they exploited the GoEmotions dataset \citep{demszky-etal-2020-goemotions} for capturing emotion information. With regards to the first method, the researchers defined this approach as alternating multi-task models, since these models are simply two single-task models sharing the same base BERT representation layers. With regards to the second approach, the authors introduced a traditional multi-task learning architecture, in which the two tasks are learned simultaneously using the same input. Finally, the third approach corresponds to a fine-tuning method. A different multi-task learning framework was introduced by \citet{yadav-etal-2020-identifying} for identifying depression in Twitter, where the figurative usage detection constitutes the auxiliary task. Specifically, the authors predict three classes in terms of the figurative usage detection task, namely metaphor, sarcasm, and others. The primary task is the symptom identification with nine labels from PHQ-9. A multi-task learning framework was also introduced by \citet{8910655}, where the tasks correspond to the regression and classification of the level of depression based on the PHQ-8 score. Specifically, the authors descretized the PHQ-8 score for the classification task, while the PHQ-8 scores were directly predicted for the regression task. The authors exploited acoustic, textual, and visual modalities for training the introduced architecture. A similar approach was introduced by \citet{9141494}, where the authors defined three tasks, namely the depression level regression, depression level classification, and emotion intensity regression. In terms of the depression level regression task, the authors predicted simply the PHQ-8 score, while regarding the depression level classification task the authors discretized the PHQ-8 score into five classes. In terms of the emotion intensity regression task, the authors exploited the CMU-MOSEI dataset. Then, three different multitask models were introduced. Similarly, \citet{9669546} explored the use of emotional knowledge to improve depression detection by introducing a multimodal and multi-task learning framework. The authors exploited inter-modal ($I_e$) and inter-task ($I_t$) attentions for learning multi-modal fused representation and the relationship between depression detection and emotion recognition respectively. 

A feature extraction procedure was adopted by \citet{8681445}, where the authors extracted n-grams, LIWC features, and Latent Dirichlet Allocation (LDA) topics and trained shallow machine learning classifiers. Specifically, LR, Support Vector Machine, AdaBoost, RF, and Multilayer Perceptron (MLP) were trained. Findings showed that the MLP classifier trained with LIWC, LDA, and bigrams achieved the highest classification results.

Another approach was introduced by \citet{9723579} for automatically identifying depression, anxiety, and their comorbidity. Specifically, the authors exploited a multi-label solution for representing the comorbidity condition. Additionally, the authors introduced a stacking ensemble approach consisting of two levels. On the other hand, the authors in \citet{10068655} redesigned the multilabel problem as a multi-task classification where one post indicates both depression and anxiety.

\citet{10154134} introduced a method, which injects extra linguistic information into pretrained language models based on transformers via an Attention Gating Mechanism. Specifically, the authors integrated LIWC features, top2vec, LDA topics, and NRC Sentiment Lexicon features into BERT and MentalBERT. Also, they employed label smoothing for calibrating the proposed models and evaluated their approaches in terms of both the performance and the calibration.

\citet{9658177} proposed a feature extraction approach and trained several machine learning algorithms for predicting depressive posts. Specifically, the authors extracted sentiment lexicon features, content based features, pos-tags, linguistic characteristics, and readability scores. Finally, Logistic Regression, Support Vector Machine, Decision Tree, Multilayer Perceptron, Bagging Predictors (BP), Random Forest, Adaptive Boosting (AB), and Gradient Boosting (GB) were trained using the aforementioned feature set.

The authors in \citet{info:doi/10.2196/28754} proposed a deep neural network consisting of a semantic understanding network, an emotion understanding network, and a depression detector. Specifically, the semantic understanding network aims at capturing the contextual semantic information in the text for depression detection, while the emotion understanding network aims to identify the emotional semantic information in the text for depression detection.

\subsection{Related Work Review Findings}

From the abovementioned research works, it is evident that research works having introduced multi-task learning settings, use as primary and auxiliary tasks the depression/stress and emotion tasks respectively. At the same time, the majority of these research works have exploited their approaches using the DAIC-WOZ Database \citep{10.1145/3347320.3357688,10.5555/2615731.2617415,gratch-etal-2014-distress}, which contains clinical interviews. Thus, little work has been done in terms of proposing multitask learning frameworks for identifying stress and depression via social media. To be more precise, there are limited studies having reformulated the multilabel problem, i.e., posts indicating both depression and anxiety, into a multi-task learning framework by utilizing one single dataset. No study has experimented with exploiting multiple datasets and using as primary and auxiliary tasks the depression and stress tasks respectively using social media posts. In addition, despite the rise of deep learning, many studies still employ feature extraction approaches and train shallow machine learning algorithms. 

Therefore, our work differs significantly from the research works mentioned above, since we \textit{(a)} present the first study, which exploits two different datasets, for proposing a multi-task learning framework, where we define the depression detection task as the main task and the stress detection task as the auxiliary task, for predicting depressive and stressful posts in social media, \textit{(b)} we introduce the attention fusion network in the multi-task learning framework, and \textit{(c)} compare our approaches with existing research initiatives, single-task learning architectures (by stacking two BERT models), and transfer learning strategies.


\section{Datasets}

In this section, we describe the datasets used for conducting our experiments. Specifically, we experiment with depression and stress datasets.

\noindent \textbf{Stress Detection Task:} We use the Dreaddit dataset introduced in \citet{turcan-mckeown-2019-dreaddit}, which includes stressful posts in Reddit belonging to five categories. This dataset includes 3553 posts. Specifically, this dataset includes five domains of stressful posts, namely abuse, social,  anxiety, Post-Traumatic Stress Disorder (PTSD), and financial domain. This dataset has been annotated by human annotators using the Amazon Mechanical Turk. Specifically, each worker was asked to label five posts as "Stress", "Not Stress", or "Can't Tell". Each post has been labelled by at least five workers. Finally, the label of each post was defined as the annotators' majority vote.

\noindent \textbf{Depression Detection Task:} We use the dataset introduced in \citet{pirina-coltekin-2018-identifying}. This dataset consists of 2822 posts. This dataset includes posts both from Reddit and English depression forums \citep{Ramirez-Esparza_Chung_Kacewic_Pennebaker_2021}. In terms of english depression forums \citep{Ramirez-Esparza_Chung_Kacewic_Pennebaker_2021}, the authors exploited the breast cancer group and collected one single post from a screen name for avoiding multiple entries from a single user. Additionally, the authors verified that the writer of each post was a woman, while they discarded posts, which explicitly mentioned that the writer was experiencing depression. For labelling depressive posts, the authors in \citet{pirina-coltekin-2018-identifying} adopt a protocol similar to \cite{coppersmith-etal-2015-clpsych,yates-etal-2017-depression} and are searching for expressions like "I was just diagnosed with depression" on the Depression subreddit. In terms of the non-depressive posts, the authors collect sets of posts belonging to the breast discussion subreddit, family, and friendship advice subreddits.

\section{Approach}

\subsection{Single-Task Learning}

We use Single-Task Learning (STL) models, which exploit stress and depression detection tasks as the sole optimization objectives.  The STL experiments are conducted for each primary-task dataset separately. Specifically, each post is passed through two stacked BERT layers followed by a dense layer consisting of two units for getting the final prediction. BERT is a language model based on transformers \citep{10.5555/3295222.3295349} and is pretrained on large corpora. It learns language representations by jointly conditioning on both left and right context in all layers. BERT is trained for masked language modelling, where some percentage of the input
tokens is masked at random, and then the task is to predict those masked tokens given the context. BERT has shown increased performance in many tasks, including dementia detection through spontaneous speech \citep{9769980}, complaint identification in social media \citep{jin-aletras-2020-complaint}, detection of parody tweets \citep{maronikolakis-etal-2020-analyzing}, and so on.

Formally, let $p=[p_1, p_2, ..., p_n]$ be the depressive or the stressful post. We use the BERT tokenizer and pad each post/text to a maximum length of 512 tokens. BERT tokenizer returns the input\_ids and the attention mask per post, which will be given as input to the BERT model.

Next, each post is passed through a BERT \citep{devlin-etal-2019-bert} model. Let $z \in \mathbb{R}^{N \times d}$ be the output of the BERT model. $N$ denotes the sequence length and is equal to 512, while $d$ indicates the dimensionality of the model and is equal to 768.

Next, we pass $z$ and the attention mask through the second BERT model. To do this, we set \textit{input\_ids} to None, while we use $z$ as the input embeddings. We get the [CLS] token of the BERT model as the output. Finally, we use a dense layer consisting of two units for getting the final prediction. 

The network parameters are optimized to minimize the cross entropy loss function.

\subsection{Our Proposed Multi-Task Approaches}

According to \citet{10.5555/3091529.3091535,caruana1997multitask}, it is beneficial to learn several tasks simultaneously than to learn the same tasks separately, since information and features can be shared between the tasks. In this section, we describe our proposed multi-task learning architectures for detecting stress and depression.

\subsubsection{Double Encoders}

\begin{figure*}[!htb]
    \centering
    \includegraphics[width=\textwidth]{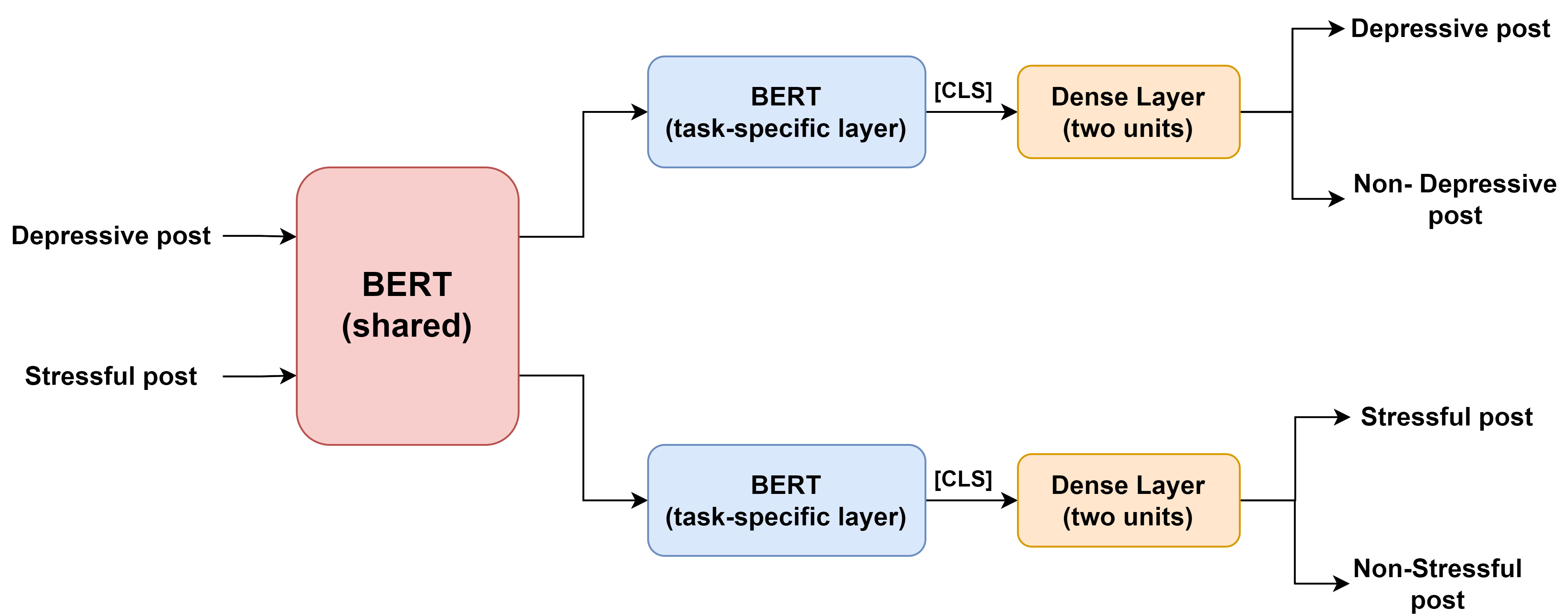}
    \caption{Our proposed MTL approach consisting of two BERT encoders}
    \label{fig:my_label}
\end{figure*}

 Our proposed architecture is illustrated in Fig.~\ref{fig:my_label}.

Let $p=[p_1, p_2, ..., p_n]$ and $s=[s_1, s_2, ..., s_T]$ be the depressive and the stressful posts respectively. We use the BERT tokenizer and pad each post/text to a maximum length of 512 tokens. BERT tokenizer returns the input\_ids and the attention mask per post, which will be given as input to the BERT model. 

Next, as illustrated in Fig.~\ref{fig:my_label}, each depressive and stressful post are passed through a shared BERT \citep{devlin-etal-2019-bert} model. Let $z \in \mathbb{R}^{N \times d}$ be the output of the BERT model corresponding to the depressive post. Similarly, let $s \in \mathbb{R}^{N \times d}$ be the output of the BERT model corresponding to the stressful post. $N$ denotes the sequence length and is equal to 512, while $d$ indicates the dimensionality of the model and is equal to 768.

Next, we use two task-specific encoders. Specifically, we exploit two BERT models, which are updated by each task separately. Formally, we pass $z$ and the attention mask corresponding to the depressive post through the BERT model. Similarly, we pass $s$ and the attention mask corresponding to the stressful text through an independent BERT model. We get the [CLS] token of each BERT model as the output. Finally, we use dense layers consisting of two units for getting the final prediction. Specifically, we use a dense layer consisting of two units for predicting depressive posts, while we use a dense layer with two units for recognizing stressful posts.

\subsubsection{Attention-Fusion Network}

\begin{figure*}
    \centering
    \includegraphics[width=\textwidth]{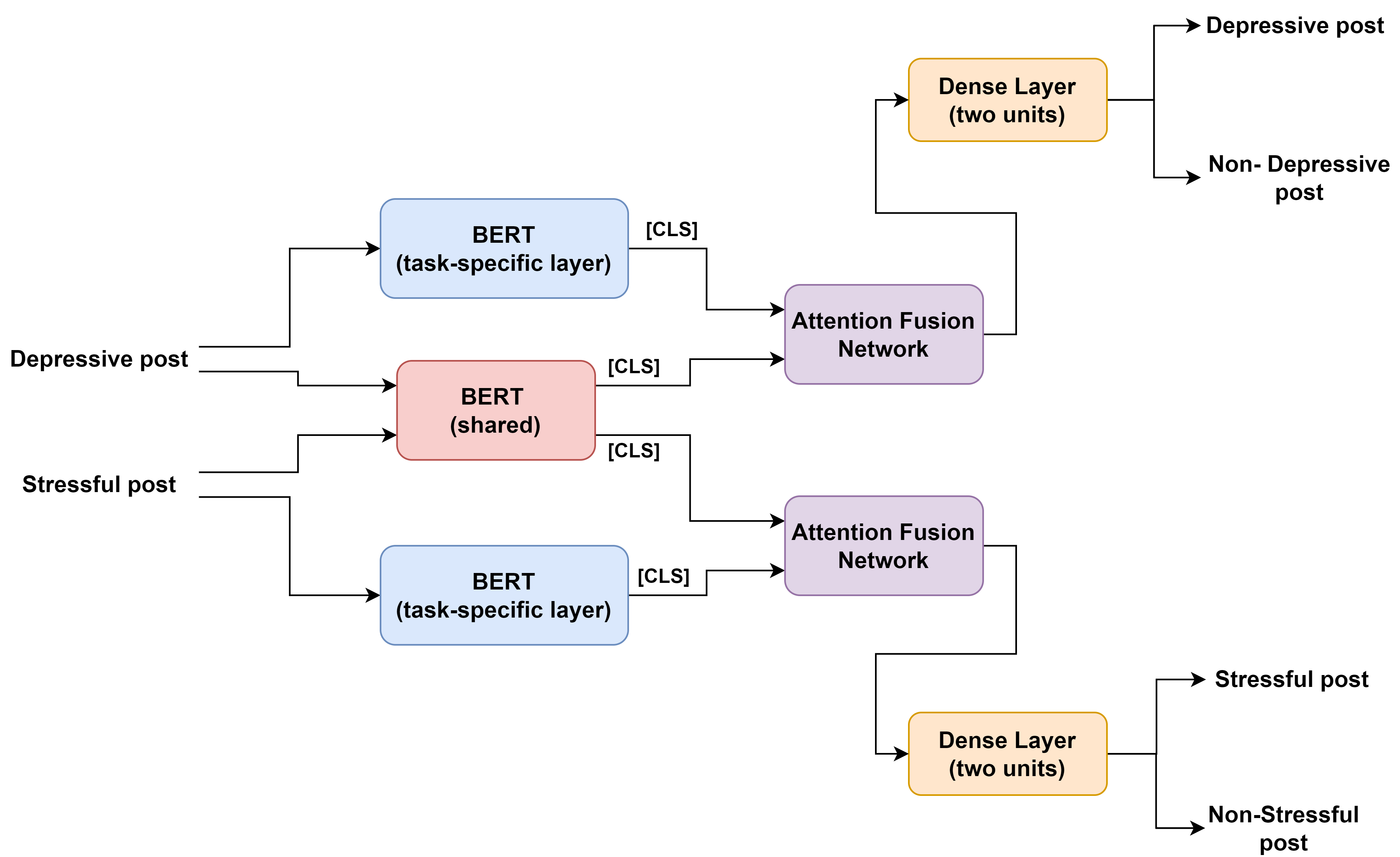}
    \caption{Our proposed MTL architecture including an Attention Fusion Network}
    \label{fig:my_label_2}
\end{figure*}

Our proposed architecture is illustrated in Fig.~\ref{fig:my_label_2}.

Let $p=[p_1, p_2, ..., p_n]$ denote the depressive post. We use the BERT tokenizer and pad each post/text to a maximum length of 512 tokens. BERT tokenizer returns the input\_ids and the attention mask per post, which will be given as input to the BERT model. 

Next, as illustrated in Fig.~\ref{fig:my_label_2}, each depressive post is passed through a shared BERT \citep{devlin-etal-2019-bert} model. We get the [CLS] token as the output of the shared BERT model.

Also, we use task-specific BERT layers. In terms of the depression recognition task, we pass each depressive post through a task-specific BERT model and get the [CLS] token as the output.

Then, the outputs of the task-specific layer and the shared layer are passed through an attention fusion network. Motivated by \citet{9141494, 8215597}, we design an attention fusion network, where we concatenate the representation vectors obtained by task-specific and shared layer and pass this vector to two ReLU activated dense layers consisting of 768 and 128 units. Finally, we use a dense layer consisting of two units with a softmax activation function and receive two values, namely $\alpha_{task}$ and $\alpha_{shared}$. These values  weight the task-specific BERT model and the shared BERT model, respectively, in computing the final vector representation. Next, we multiply $\alpha_{task}$ with the output of the task-specific BERT layer. Similarly, we multiply $\alpha_{shared}$ with the output of the shared BERT layer. Finally, we sum the corresponding products. The attention fusion network aims at understanding the behavior of each of the task-specific and shared features. For instance, if the task-specific embeddings are more important than the embeddings extracted by the shared layer, then $\alpha_{task}$ would be greater than 0.5, while $\alpha_{shared}$ would be less than 0.5. In this way, the deep learning model is capable of learning the importance of the shared and task-specific embeddings by itself in order to make the final prediction.

Next, the output of the attention fusion network is passed through a dense layer consisting of two units for getting the final prediction.

Similarly, we adopt the same procedure in terms of the stress detection task.

\subsubsection{Loss Function}

The two tasks are learned jointly by minimizing the following loss function:

\begin{equation}
L= \left(1 - \beta \right)L_{depression} + \beta L_{stress}
\label{equation3}
\end{equation}
,where $L_{depression}$ and $L_{stress}$ are the losses of depression and stress identification respectively. Specifically, the cross-entropy loss is suitable for both depression and stress detection tasks. Therefore, $L_{depression}$ and $L_{stress}$ correspond to the cross-entropy loss function. $\beta$ is a hyperparameter controlling the importance of each task. 

\section{Experiments}

\subsection{Baselines} \label{baselines}

We compare our introduced approaches with the following methods:

\begin{itemize}
\item Comparison with state-of-the-art approaches
\begin{itemize}
    \item Primary Task (Depression detection task)
    \begin{itemize}
        \item BERT: We report the performance of BERT obtained by \citet{YANG2022102961}.
        \item M-BERT (top2vec) \citep{10154134}: This method injects top2vec features into a BERT model via an Attention Gating mechanism.
        \item LR+Features \citep{8681445}: This method extracts linguistic features, including LIWC features, n-gram features, etc., and trains a Logistic Regression Classifier. We report the results reported in \cite{YANG2022102961}.
        \item BiLSTM\_Att \citep{info:doi/10.2196/28754}: This method exploits a BiLSTM coupled with an Attention mechanism. We report the results reported in \cite{YANG2022102961}.
        \item CNN \citep{husseini-orabi-etal-2018-deep}: This approach applies three convolutions each of which has 128 features and filters of the lengths 3, 4, and 5. Next, a max-pooling layer is used. Finally, the feature representations are concatenated into a single output. We report the results reported in \cite{YANG2022102961}. 
    \end{itemize}
\end{itemize}

\begin{itemize}
    \item Auxiliary Task (Stress detection task)
    \begin{itemize}
        \item BERT: We report the performance of BERT obtained by \citet{turcan-mckeown-2019-dreaddit}.
        \item M-BERT (LIWC) \citep{10154134}: This method injects LIWC features into a BERT model via an Attention Gating mechanism.
        \item KC-Net \citep{YANG2022102961}: This method consists of a context-aware post encoder, knowledge-aware dot product attention, and supervised contrastive learning.
        \item KC-Net+RoBERTa \citep{YANG2022102961}: This method is based on KC-Net and exploits a pretrained RoBERTa model.
        \item EMO\_INF \citep{turcan-etal-2021-emotion}: This method employs a multitask learning framework, in which the main task corresponds to the stress detection task, while the auxiliary task is defined as an emotion prediction task. We report the results reported in \cite{YANG2022102961}. 
        \item LR+Features \citep{8681445}: This method extracts linguistic features, including LIWC features, n-gram features, etc., and trains a Logistic Regression Classifier. We report the results reported in \cite{YANG2022102961}.
    \end{itemize}
\end{itemize}
\end{itemize}

\begin{itemize}
    \item Transfer Learning: We first train the STL model on the stress detection task until convergence and then fine-tune it for the depression task and vice versa.
\end{itemize}

\subsection{Experimental Setup}

We divide the datasets into a train, validation, and test set (70-10-20\%). Each task has its own optimizer. Specifically, the Adam optimizer is used with a learning rate of 1e-5 for both tasks. We apply \textit{EarlyStopping} based on the validation loss and stop training if the loss has not presented a decline after 8 consecutive epochs. Also, we use \textit{StepLR} with a step size of 5 and a gamma of 0.1. We use a batch size of 4. We train our introduced models for a maximum of 15 epochs, choose the epoch with the smallest validation loss, and test the model on the test set. For Double Encoders, we set $\beta$ of Eq.~\ref{equation3} equal to 0.01. For Attention Fusion Network, we set $\beta$ of Eq.~\ref{equation3} equal to 0.1. In terms of the Transfer Learning approach, we first train our STL model on the depression (stress) task with a learning rate of 1e-4, Adam optimizer, StepLR, EarlyStopping for a maximum of 15 epochs, and fine-tune it on the stress (depression) task with a learning rate of 1e-5. We use the BERT base uncased version via the Python library, namely Transformers \citep{wolf-etal-2020-transformers}. We use PyTorch \citep{NEURIPS2019_bdbca288} and train our experiments on a single Tesla P100-PCIE-16GB GPU. 

\subsection{Evaluation Metrics}

We evaluate the performance of our models by using Precision, Recall, F1-score, Accuracy, and Specificity. We compute these metrics by considering the stressful/depressive posts as the positive class (label: 1), while the non-stressful / non-depressive posts correspond to the negative class (label: 0).

\section{Results}

The results of our introduced approaches are reported in Tables \ref{primary_task} and \ref{auxiliary_task}. Specifically, Table~\ref{primary_task} reports the results on the primary task, i.e., depression detection task, while Table~\ref{auxiliary_task} reports the results on the auxiliary task, i.e., stress detection task. Also, these tables present a comparison of our approaches with baselines described in Section \ref{baselines}.

\begin{table*}[!htb]
\centering
\caption{Performance comparison among proposed models and state-of-the-art approaches on the primary task (depression detection task).}
\begin{tabular}{lccccc}
\toprule
\multicolumn{1}{l}{}&\multicolumn{5}{c}{\textbf{Evaluation metrics}}\\
\cline{2-6} 
\multicolumn{1}{l}{\textbf{Architecture}}&\textbf{Precision}&\textbf{Recall}&\textbf{F1-score}&\textbf{Accuracy}&\textbf{Specificity}\\
\midrule
\multicolumn{6}{>{\columncolor[gray]{.8}}l}{\textbf{Comparison with state-of-the-art approaches}} \\
BERT \citep{YANG2022102961} & 91.40 & 91.40 & 91.40 & - & -  \\
M-BERT (top2vec) \citep{10154134} & 90.34 & 94.93 & 92.58 & 92.57 & - \\
LR+Features \citep{8681445} & 89.00 & 92.00 & 89.00 & - & - \\
BiLSTM\_Att \citep{info:doi/10.2196/28754} & 90.40 & 95.00 & 92.60 & - & - \\
CNN \citep{husseini-orabi-etal-2018-deep} & 85.20 & 85.10 & 85.10 & - & - \\
\midrule
\multicolumn{6}{>{\columncolor[gray]{.8}}l}{\textbf{Transfer Learning}} \\
 & 89.97 & 95.05 & 92.44 & 92.21 & 89.36 \\ \midrule
\multicolumn{6}{>{\columncolor[gray]{.8}}l}{\textbf{Single-Task Learning}} \\
 & 94.31 & 90.75 & 92.49 & 92.39 & 94.14 \\ \midrule
\multicolumn{6}{>{\columncolor[gray]{.8}}l}{\textbf{Multi-Task Learning}} \\
Double Encoders & 94.01 & 91.94 & 92.96 & 93.27 & 94.52 \\
Attention Fusion Network & 93.57 & 90.31 & 91.91 & 92.73 & 94.77 \\
\bottomrule
\end{tabular}
\label{primary_task}
\end{table*}

\begin{table*}[!hbt]
\centering
\caption{Performance comparison among proposed models and state-of-the-art approaches on the auxiliary task (stress detection task).}
\begin{tabular}{lccccc}
\toprule
\multicolumn{1}{l}{}&\multicolumn{5}{c}{\textbf{Evaluation metrics}}\\
\cline{2-6} 
\multicolumn{1}{l}{\textbf{Architecture}}&\textbf{Precision}&\textbf{Recall}&\textbf{F1-score}&\textbf{Accuracy}&\textbf{Specificity}\\
\midrule
\multicolumn{6}{>{\columncolor[gray]{.8}}l}{\textbf{Comparison with state-of-the-art approaches}} \\
BERT \citep{turcan-mckeown-2019-dreaddit} & 75.18 & 86.99 & 80.65 & - & -  \\
M-BERT (LIWC) \citep{10154134} & 77.21 & 89.97 & 83.10 & 81.12 & - \\
KC-Net \citep{YANG2022102961} & 84.10 & 83.30 & 83.50 & - & - \\
KC-Net+RoBERTa \citep{YANG2022102961} & 82.70 & 82.60 & 82.70 & - & - \\
EMO\_INF \citep{turcan-etal-2021-emotion} & 81.70 & 81.70 & 81.70 & - & - \\
LR+Features \citep{8681445} & 73.50 & 81.00 & 77.00 & - & - \\
\midrule
\multicolumn{6}{>{\columncolor[gray]{.8}}l}{\textbf{Transfer Learning}} \\
 & 82.71 & 85.02 & 83.85 & 83.37 & 81.65 \\ \midrule
\multicolumn{6}{>{\columncolor[gray]{.8}}l}{\textbf{Single-Task Learning}} \\
 & 81.73 & 86.44 & 84.02 & 82.83 & 78.88 \\ \midrule
\multicolumn{6}{>{\columncolor[gray]{.8}}l}{\textbf{Multi-Task Learning}} \\
Double Encoders & 77.58 & 86.79 & 81.93 & 79.47 & 70.99 \\
Attention Fusion Network & 82.24 & 85.03 & 83.61 & 82.66 & 80.00 \\
\bottomrule
\end{tabular}
\label{auxiliary_task}
\end{table*}

As one can easily observe in Table~\ref{primary_task}, our introduced multi-task learning architectures obtain better performance than state-of-the-art approaches, single-task learning, and transfer learning strategies. Specifically, Double Encoders constitutes our best performing model outperforming the other approaches in F1-score by 0.36-7.86\% and in Accuracy by 0.54-1.06\%. Although there are approaches obtaining better Precision and Recall scores than Double Encoders, it is worth noting that Double Encoders outperforms all these approaches in F1-score, which constitutes a weighted average of Precision and Recall. In addition, we observe that Attention Fusion Network outperforms Double Encoders in Specificity by 0.25\%. However, F1-score is a more suitable metric in health-related studies, since high Specificity and low F1-score indicates that people in stressful/depressive conditions are misdiagnosed as being in non-stressful/non-depressive conditions. Also, we observe that our proposed multi-task learning architecture outperforms transfer learning. This can be justified by a variety of reasons. Specifically, in the case of transfer learning, the primary task fine-tunes the knowledge from the auxiliary task for its task objective and therefore may be forgetting auxiliary task knowledge. In contrast, our proposed multi-task learning architecture includes both task-specific and shared layers. Therefore, each task learns some knowledge, which is proven to be beneficial to both related tasks leading to a better generalization. Finally, we observe that our proposed single-task learning approach by stacking BERT layers yields better performance than the one obtained by a single BERT layer \citep{YANG2022102961}. Specifically, our single-task learning approach outperforms BERT \citep{YANG2022102961} in F1-score by 1.09\%.

As one can easily observe in Table~\ref{auxiliary_task}, our proposed multi-task learning architectures obtain worse results than the other approaches, single-task learning and transfer learning approaches. This can be easily justified by the fact that the stress detection constitutes an auxiliary task in a multi-task learning framework. More specifically, we have set $\beta$ of Eq.~\ref{equation3} equal to 0.01 in terms of Double Encoders, while the parameter $\beta$ is equal to 0.1 in terms of the Attention Fusion Network. Therefore, we place more importance to the primary task (depression detection task) than the auxiliary one (stress detection task). However, we observe that our introduced Attention Fusion Network outperforms the state-of-the-art approaches in F1-score by 0.11-6.61\%. In addition, we observe that our introduced single-task learning model outperforms the state-of-the-art approaches in F1-score by 0.52-7.02\%. Finally, we observe that the transfer learning strategy, i.e., training on the depression detection task and fine-tuning on the stress detection task, attains better Accuracy score by 0.54\% than the single-task learning model, while it also surpasses the state-of-the-art approaches in F1-score by 0.35-6.85\%.

Overall, if one is interested in identifying depressive posts in social media, our proposed multitask learning architecture, namely Attention Fusion Network, appears to be the best performing one. Specifically, it outperforms some state-of-the-art approaches, transfer learning, and single-task learning approaches reaching an Accuracy and F1-score up to 92.73\% and 91.91\% respectively. On the other hand, if someone is interested in exploiting a model for identifying stressful posts in social media, our transfer learning approach, i.e., first training of the single-task learning deep learning model on the depression detection task until convergence and after fine-tuning it on the stress detection task, seems to be the best performing one attaining an Accuracy and F1-score of 83.37\% and 83.85\% respectively.

\subsection{Effect of hyperparameter $\beta$ on the performance}

\begin{figure*}[!htb]
\centering
\begin{subfigure}[t]{0.95\columnwidth}
\includegraphics[width=\linewidth]{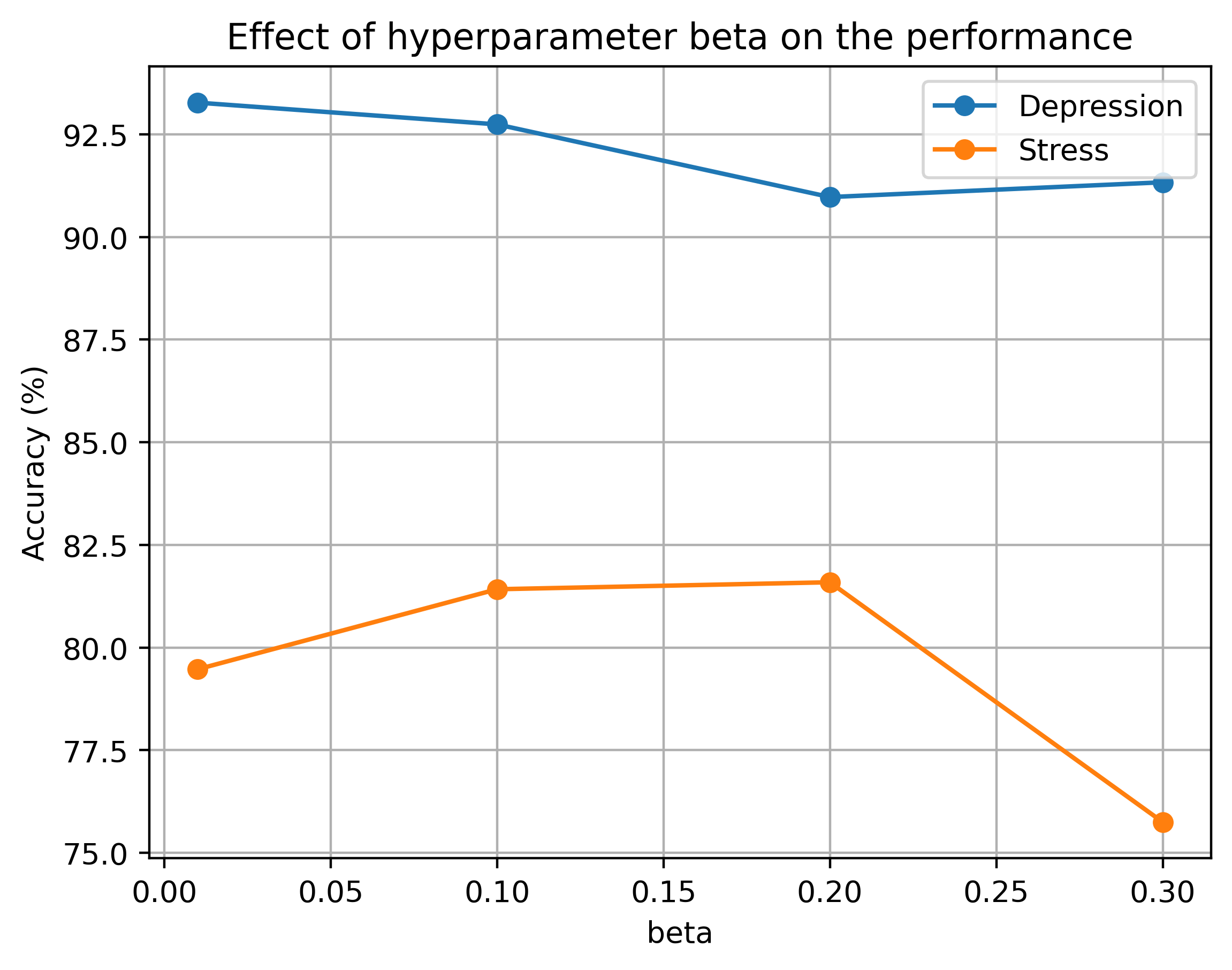}
\caption{Accuracy}
\label{cbsa_1}
\end{subfigure}
\begin{subfigure}[t]{0.95\columnwidth}
\centering
\includegraphics[width=\linewidth]{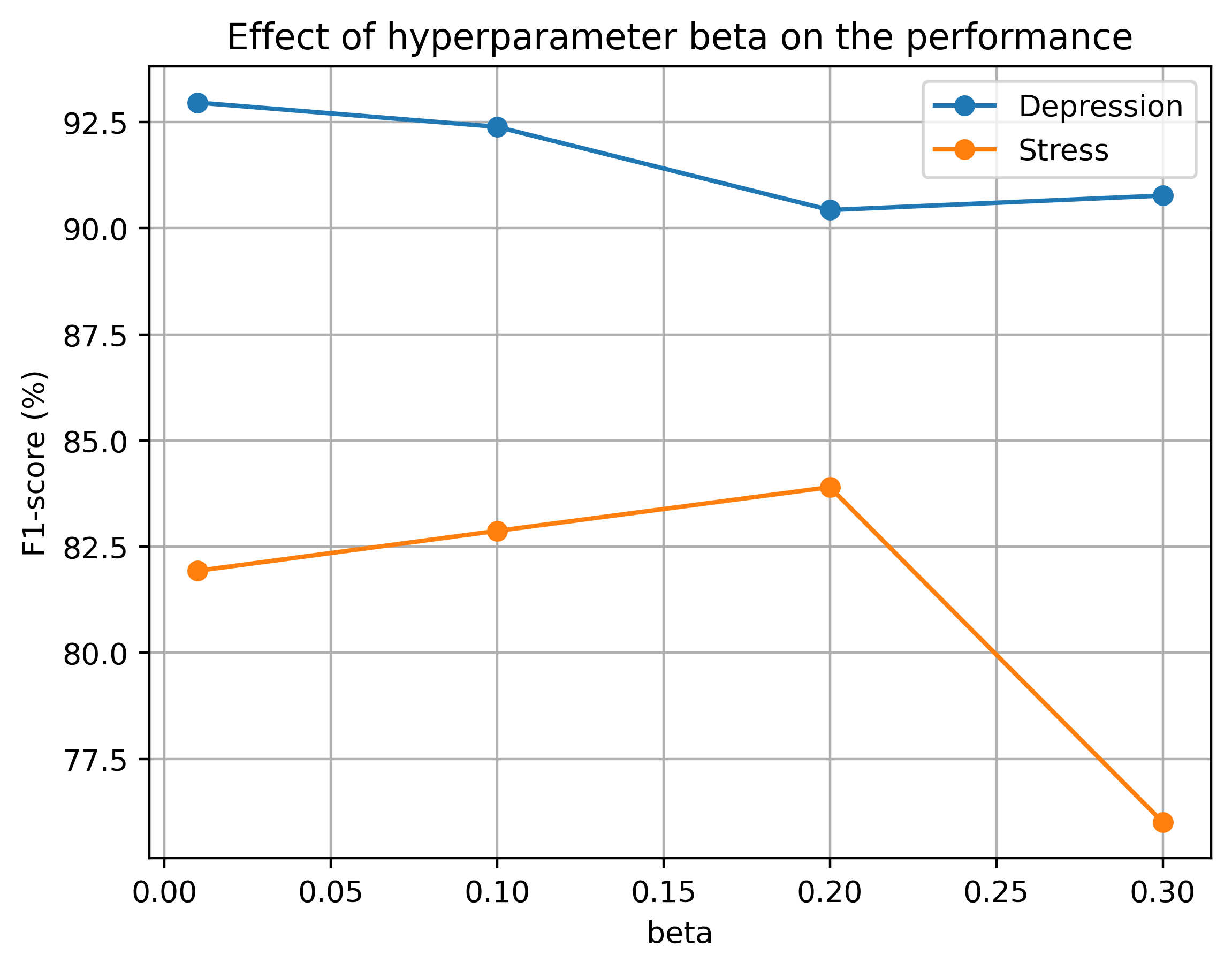}
\caption{F1-score}
\label{gated_sum_1}
\end{subfigure}
\caption{Effect of hyperparameter $\beta$ on the performance (Double Encoders)}
\label{context_based_self_attention_1}
\end{figure*}

\begin{figure*}[!htb]
\centering
\begin{subfigure}[t]{0.95\columnwidth}
\includegraphics[width=\linewidth]{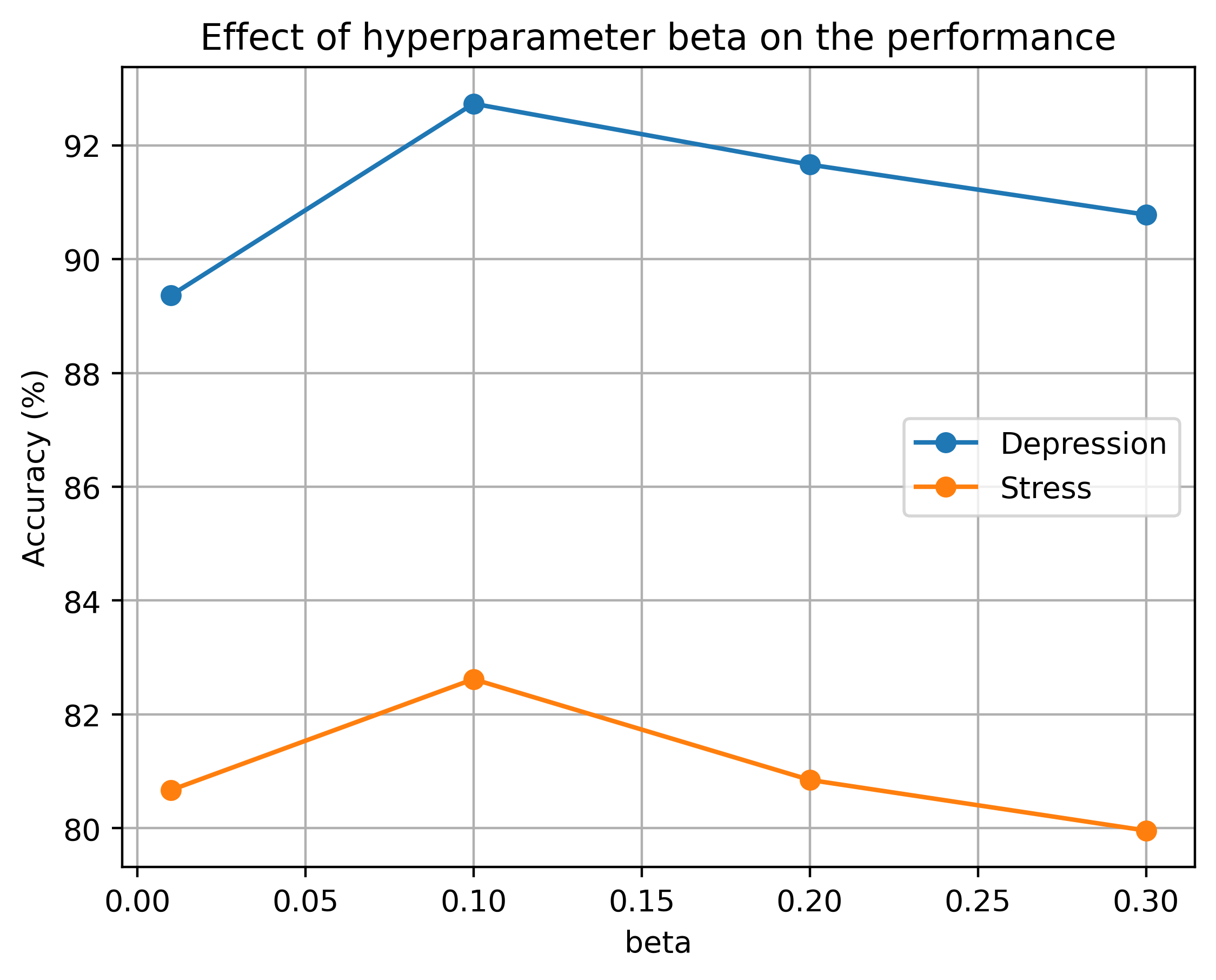}
\caption{Accuracy}
\label{cbsa}
\end{subfigure}
\begin{subfigure}[t]{0.95\columnwidth}
\centering
\includegraphics[width=\linewidth]{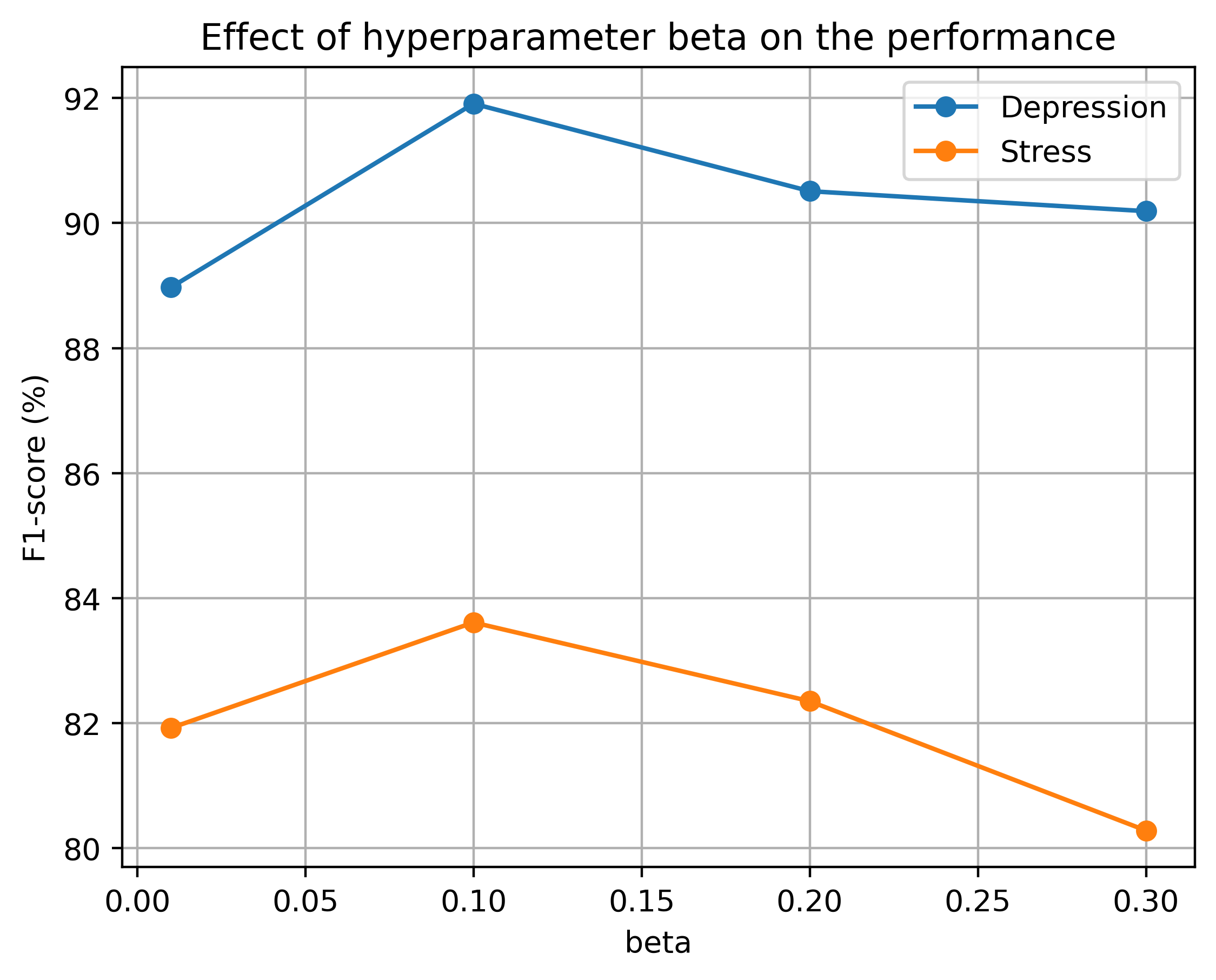}
\caption{F1-score}
\label{gated_sum}
\end{subfigure}
\caption{Effect of hyperparameter $\beta$ on the performance (Attention Fusion Network)}
\label{context_based_self_attention}
\end{figure*}

The $\beta$ parameter is a hyperparameter that controls the importance we place on each task as is shown in Eq.~\ref{equation3}. Fig.~\ref{context_based_self_attention_1} shows the Accuracy and F1-score of our proposed multi-task learning framework (Double Encoders) with different $\beta$ settings. Fig.~\ref{context_based_self_attention} shows the Accuracy and F1-score of our proposed multi-task learning framework (Attention Fusion Network) with different $\beta$ settings. The hyperparameter $\beta$ is set to \{0.01,0.1,0.2,0.3\}. 

As one can easily observe in Fig.~\ref{context_based_self_attention_1}, setting the parameter $\beta$ equal to 0.01 yields the highest Accuracy and F1-score in terms of the depression detection task (primary task). Increasing the value of this parameter to 0.10 and 0.20, we observe that both the Accuracy and F1-score present a decrease. Finally, setting the parameter $\beta$ equal to 0.30, we observe that a slightly better performance is attained than the one obtained by setting $\beta$ equal to 0.20.

In Fig.~\ref{context_based_self_attention}, we observe that similar patterns in the performance are obtained in terms of both the primary and the auxiliary task by varying the values of $\beta$. Specifically, we observe that the best Accuracy and F1-score are obtained for both the primary and the auxiliary tasks by setting $\beta$ equal to 0.10. As $\beta$ increases, both Accuracy and F1-score present a decline.

\section{Conclusion}

In this paper, we introduced two architectures to detect stressful and depressive posts in social media through a multi-task learning framework. Each multi-task learning framework consists of two tasks, namely the primary task and the auxiliary one. We defined the primary task as the depression detection task, while the stress prediction task corresponds to the auxiliary task. In terms of the first approach, each stressful and depressive post pass through a shared BERT layer, which is updated by both tasks. Then, we exploited two task-specific BERT layers for predicting stressful and depressive posts. In terms of the second approach, we exploited both shared and task-specific BERT layers. Additionally, we included an Attention Fusion Network for capturing the most important information extracted by the shared and task-specific layers. We utilized two datasets, one corresponding to stressful and non-stressful posts, and the other one corresponding to depressive and non-depressive posts. These datasets have been collected under different circumstances, thus rendering the recognition of depression and stress in a single neural network more challenging. We compared our approaches with existing research initiatives, single-task learning, and transfer learning strategies. Findings showed that MTL with stress detection is beneficial for the depression detection task. Our results also suggested the superiority of MTL over STL for depression detection.

However, this study has some limitations. Specifically, it does not include explainability techniques for rendering the proposed models explainable. Additionally, we did not perform hyperparameter tuning due to limited access to computational resources. Hyperparameter tuning often leads to performance improvement. In addition, this study requires labelled dataset. On the contrary, methods of self-supervised learning have been proposed lately for addressing the issue of labels' scarcity. 

In the future, we aim to apply explainability techniques for explaining the predictions made by our models. Additionally, we plan to exploit our introduced models in a federated learning framework. Finally, one of our future plans is to create a hybrid approach by combining the two introduced models, i.e., \textit{Double Encoders} and \textit{Attention-Fusion Network}.

\section*{Ethical Considerations}

This work uses publicly available data. No personal information was collected.

\printcredits

\bibliographystyle{cas-model2-names}

\bibliography{cas-refs}

\end{document}